\documentclass{article}

\usepackage{ijcai24}
\usepackage{amsmath}
\usepackage{amssymb}
\usepackage{amsthm}
\usepackage{times}
\usepackage{soul}
\usepackage{url}
\usepackage[hidelinks]{hyperref}
\usepackage[utf8]{inputenc}
\usepackage[small]{caption}
\usepackage{graphicx}
\usepackage{bm}
\usepackage{booktabs}
\usepackage{algorithm}
\usepackage{algorithmic}
\usepackage[switch]{lineno}

\title{The Impact of an XAI-Augmented Approach on Binary Classification with Scarce Data}

\author{
Ximing Wen$^1$,
Rosina O. Weber$^1$,
Anik Sen$^1$,
Darryl Hannan $^1$,
Steven C. Nesbit  $^1$,
Vincent Chan  $^2$,
Alberto Goffi $^2$,
Michael Morris $^3$,
John C. Hunninghake $^3$,
Nicholas E. Villalobos $^3$,
Edward Kim $^1$, 
Christopher J. MacLellan $^4$\\
\affiliations
$^1$Drexel University  \ \ \ \  \ \ \ $^2$University of Toronto \\
$^3$Brooke Army Medical Center \ \ \ \ \ \ \  $^4$Georgia Institute of Technology\\
\emails
\{xw384, rw37, as5867, dwh48, scn43\}@drexel.edu,
Vincent.Chan@uhn.ca,
Alberto.Goffi@unityhealth.to,
\{michael.j.morris34.civ, john.c.hunninghake.mil, nicholas.e.villalobos.mil\}@health.mil,
ek826@drexel.edu,
cmaclellan3@gatech.edu
}

\begin{document}

\maketitle

\begin{abstract}
    Point-of-Care Ultrasound (POCUS) is the practice of clinicians conducting and interpreting ultrasound scans right at the patient's bedside. However, the expertise needed to interpret these images is considerable and may not always be present in emergency situations. This reality makes algorithms such as machine learning classifiers extremely valuable to augment human decisions. POCUS devices are becoming available at a reasonable cost in the size of a mobile phone. The challenge of turning POCUS devices into life-saving tools is that interpretation of ultrasound images requires specialist training and experience. Unfortunately, the difficulty to obtain positive training images represents an important obstacle to building efficient and accurate classifiers. Hence, the problem we try to investigate is how to explore strategies to increase accuracy of classifiers trained with scarce data. We hypothesize that training with a few data instances may not suffice for classifiers to generalize causing them to overfit. Our approach uses an Explainable AI-Augmented approach to help the algorithm learn more from less and potentially help the classifier better generalize.
\end{abstract}

\section{Introduction}\label{sec:introduction}
In radiology imaging, the scarcity of positive data (\emph{i.e.}, illustrating each specific disease) poses a significant challenge for the development of effective diagnostic tools. Radiology relies heavily on machine learning (ML) algorithms to accurately detect and diagnose various conditions. However, without sufficient positive data, the classifiers struggle to learn the distinct features of each condition, leading to reduced accuracy and reliability.

 Researchers investigate various approaches to address the problem of limited training data in ML such as data augmentation \cite{perez2017effectiveness,heidari2020improving}, single-class learning \cite{denis1998pac,de1999positive,liu2003building}, zero-, one-, or few-shot learning \cite{larochelle2008zero,lampert2013attribute,pourpanah2022review,fei2006one,Scheirer2012toward}, regularization \cite{nowlan2018simplifying,srivastava2014dropout,nusrat2018comparison}, and with \emph{ad hoc} deep learning architectures \cite{xu2023high}. Within the field of eXplainable Artificial Intelligence (XAI), recent works focus on constructing priors that incorporate human intuition or domain knowledge to improve models' robustness and accuracy when training data is limited \cite{perez2017effectiveness,heidari2020improving}. 
 However, in their approach, usually either the interpretability of features is required ({\it i.e.} image pixels) or additional domain knowledge is used as input, which limits their applicability in occasions that the validity of features is not easily accessed. 
 
 To solve this challenge, we propose a new generalizable prior constructed with additive feature attribution \cite{erion2021improving} without any human input or domain knowledge and then demonstrate training classifiers with the prior can provide additional feedback for back-propagation thus improving classification accuracy.

 Additive feature attribution methods \cite{hastie2017generalized,lundberg2017unified} produce as output a model of the data model they aim to explain as a sum of the contributions of the individual features from data. They are originally designed to assist ML classifiers in explaining their decisions and is studied in the sub-fields of XAI (henceforth we use XAI as a general term).  These additive models have the property of {\it local accuracy}, dictating that the sum of feature attributions for a particular instance indicates its class. Consequently, {\it local accuracy} determines whether the class indicated by the explanation model is consistent with the classifier's prediction. In the ideal situation, both the decision boundary built by the explanation model and the prediction model should be similar to the ground truth data. Motivated by this, we create a new prior by constructing a cross-entropy loss between the true label and the class indicated by the explanation model utilizing Gradient SHAP \cite{erion2021improving}, an additive feature attribution method. 

The datasets used in the experiments herein illustrate three medical conditions: pneumothorax (PTX) (\emph{i.e.}, collapsed lung), intracranial pressure via optic nerve sheath diameter (ONSD), and COVID-19. 

The contributions of our work are:
  \begin{enumerate}
   \item Propose a novel approach to augment classification with scarce data with Gradient SHAP \cite{erion2021improving}.    
   \item Demonstrate the performance of the proposed approach in reducing overfitting to help improve classification performance in three datasets. 
 \end{enumerate}

The rest of this paper is organized as follows. Section \ref{sec:background} highlights the background of the baseline classifiers and provides related work. We next describe our motivation in Section \ref{sec:motivation}. Then, we introduce the XAI-Augmented approach in Section \ref{XAIA}. We conduct studies in Section \ref{sec:studies} and present results and discussion in Section \ref{sec:results}. Conclusions are summarized in Section \ref{sec:conclusion}, and limitation and future work are discussed in Section \ref{sec:limitation}.

\section{Background and Related Work}\label{sec:background}
In this section, we describe the baseline classifiers and the related works.

\subsection{The Baseline Classifiers}\label{base}
The baseline classifiers we utilize in the experiments herein are three variations of the architecture described in \citeauthor{Hannan_Nesbit_Wen_Smith_Zhang_Goffi_Chan_Morris_Hunninghake_Villalobos_Kim_Weber_MacLellan_2023} \shortcite{Hannan_Nesbit_Wen_Smith_Zhang_Goffi_Chan_Morris_Hunninghake_Villalobos_Kim_Weber_MacLellan_2023}. To the best of our knowledge, the architecture presented in \citeauthor{Hannan_Nesbit_Wen_Smith_Zhang_Goffi_Chan_Morris_Hunninghake_Villalobos_Kim_Weber_MacLellan_2023} \shortcite{Hannan_Nesbit_Wen_Smith_Zhang_Goffi_Chan_Morris_Hunninghake_Villalobos_Kim_Weber_MacLellan_2023} achieves the highest accuracy on the datasets used and thus we choose them as the baseline classifiers. Each variation is used for a different dataset, namely, PTX, ONSD, and COVID-19, adapting to their characteristics. Common to all is the use of a sparse coding model and a small data classifier with convolutional layers.

The task for PTX and COVID-19 is to classify whether a pleural line is moving shown in lung ultrasound videos. A pleural line is a terminology in radiology, which indicates where the lung comes into contact with the chest wall. Its movement is the most important feature for PTX diagnosis. We will provide more detail about the datasets in Section \ref{sec:studies}. The baseline model for PTX \cite{Hannan_Nesbit_Wen_Smith_Zhang_Goffi_Chan_Morris_Hunninghake_Villalobos_Kim_Weber_MacLellan_2023} has three components. First, a YOLOv4 object detection model \cite{bochkovskiy2020yolov4} recognizes the region of interest around the pleural line and creates a bounding box around it. Second, a 3D convolutional sparse coding model \cite{olshausen1997sparse} is learned with a convolutional variant of the Locally Competitive Algorithm (LCA) \cite{paiton2019analysis} to compute sparse features with an activation map. The representation of the activated sparse features is used in a convolutional neural network (CNN) classifier with two convolutional layers and two feed-forward layers with dropout. These classifiers are then trained with a binary cross-entropy loss (BCE) function. The classifiers for PTX and COVID-19 are designed to classify frames rather than videos. The process is to extract frames by striding over the video frames at a fixed interval. At each point, the model extracts the given frame along with the two previous and two subsequent frames, resulting in a five-frame block. When applying this binary classifier to the PTX dataset, it determines whether a frame block represents lung sliding. Sliding indicates normal and is thus negative for pneumothorax. This pipeline utilizes the output logits as confidence values, averaging these and rounding to the closest prediction.

The classifier for the COVID-19 data uses the same 3D convolutional sparse coding model as for PTX. The difference is that YOLO is not required because there are no object labels to be learned. The same small-data convolutional classifier is used.

The ONSD dataset consists of ultrasound videos of the optic nerve sheath. The goal of this task is to detect elevated intracranial pressure (ICP) by measuring the optic nerve sheath diameter (ONSD). The model is different from the PTX model in \cite{Hannan_Nesbit_Wen_Smith_Zhang_Goffi_Chan_Morris_Hunninghake_Villalobos_Kim_Weber_MacLellan_2023} in a few aspects. The YOLO is run to detect both the eye and the nerve. A 300$\times$100 region is cropped from the nerve, where the top of the region falls 3mm below the bottom of the eye and is centered on the middle of the nerve. This is then downsized to 150$\times$50 before being fed to a 2D sparse coding model. Once this region is acquired, the task has two steps: detecting the boundaries of the nerve and measuring the distance between the boundaries. The sparse coding model is constructed with both of these steps in mind, where the goal is to construct an activation map for a given image where the distance between the nerve boundaries is obvious, reducing the complexity of the task for the classifier. A standard 2D convolutional sparse coding approach would use a small filter size, such as 8$\times$8, and stride these filters over both the $x$ and $y$ dimensions to build an activation map (see PTX architecture in \citeauthor{Hannan_Nesbit_Wen_Smith_Zhang_Goffi_Chan_Morris_Hunninghake_Villalobos_Kim_Weber_MacLellan_2023} \shortcite{Hannan_Nesbit_Wen_Smith_Zhang_Goffi_Chan_Morris_Hunninghake_Villalobos_Kim_Weber_MacLellan_2023}). While this approach may yield vertical edge detectors that activate on the nerve boundaries, these filters would activate for other vertical edges as well. Additionally, striding over both the $x$ and $y$ dimensions is computationally expensive, where the cost scales quadratically with image size. Therefore, a 2D sparse coding model is trained on these regions with a filter size of 150$\times$10. This filter covers the entire width of the nerve region, resulting in the filters only sliding across the y-axis. The idea is that the edges of the nerve will still need to be represented in the sparse model and therefore it will learn filters that activate on both edges of a given nerve within the same filter. For nerves of different widths, different filters would be learned with the same overall pattern but the edges would be closer or farther based on the observed width of the nerve. The resulting sparse coding model produces activation maps that are 1$\times$M$\times$N, where N is the number of filters and M is the vertical dimension. They sum these activation maps over the vertical dimension to learn which of our learned, specialized nerve-width filters are most activated in a given image. If one of these filters fits the given nerve well, then it will be highly activated over the entire vertical dimension of the image. This results in a vector of size N, where each value corresponds to the activity level of a filter. They then learn a simple two-layer MLP that takes the N-dimensional vector as input and produces a single binary class prediction, corresponding to whether the width of the nerve exceeds the target threshold. 

\subsection{Related Work}
\citeauthor{erion2021improving}  \shortcite{erion2021improving} introduced the formal expression of attribution priors as constraints imposed when training a classifier originating from feature attribution methods. They experimented on three attribution priors: pixel attribution prior, graph attribution prior, and sparse prior. The first two penalize feature attribution differences between nearby pixels or similar features and the last one encourages a small number of feature attribution to have a large portion of total attribution. Their work shows promising results in improving models' accuracy and robustness. Another relevant work is in \citeauthor{weber2022beyond} \shortcite{weber2022beyond}. They used an attention mask to emphasize important features showing how a concept from XAI can reduce overfitting. The experiment results show that XAI-Augmented models are, on average, able to generalize better than baseline models. However, the limitation is that those studies were conducted using simplistic data with only five features. 

\citeauthor{barnett2021case} \shortcite{barnett2021case} leveraged GradCAM's \cite{selvaraju2017grad} saliency map in a model of deep case-based reasoning \cite{chen2019looks} where prototypes are learned to support a similarity-based classification. The model is interpretable because the visible prototypes are the reasons for the classification. When implementing the same approach in breast cancer X-rays, they aligned their model with experts' annotations incorporated into the loss. The result is an improvement in the model's accuracy and better alignment of the explaining prototypes with domain knowledge. Similar work by \citeauthor{liu2019incorporating} \shortcite{liu2019incorporating} used Integrated Gradients \cite{erion2021improving} to incorporate priors into the loss function via an XAI-based regularization term in text classification. Their work also shows improvement in accuracy. However, both of their work involves great effort from human experts to annotate images. These annotations serve as the validity of the generated explanation and play a crucial role in improving the model's performance. In lots of domains, the validity of attribution values cannot be easily assessed due to the high cost of human annotations or for not being interpretable by humans but needs additional human effort.

Human involvement is completely left out in \citeauthor{ross2018improving} \shortcite{ross2018improving}, where authors improved the model's robustness and interpretability by regularizing their input gradients, although the authors note that the model's decision boundaries change. In our approach, we make use of the local accuracy and additive attribution of Gradient SHAP to construct attribution prior without any additional human input. To the best of our knowledge, these are the approaches that most resemble ours and none has used the same method as we did or used such methods with real-world ultrasound videos to analyze the impact of this technology.

\section{Motivation to Use XAI to Augment Training}\label{sec:motivation}
Given small data classifiers $f$, we have input features $x \in X$, $X \subset \mathbb{R}^d$ and binary labels $y \in Y, Y=\{0,1\}$ ({\it i.e}., $y = 0$ when class is negative for disease and $y = 1$ when class is positive for disease). These data represent the real-world data, which characterizes a ground-truth decision boundary that separates the classes 1 and 0. Using the set of $n$ training instances $x$, we train a classifier $f(x)$ that learns to assign labels ${f(x)}$ for each instance. By producing these values for ${f(x)}$, $f(x)$ proposes its characterization of the decision boundary. In practice with the datasets studied herein, $f(x)$ obtains an accuracy level describing the proportion of correct classifications over the total predictions $a_f \neq 1$ hence producing an error $E_f \neq 0$. Consequently, its decision boundary does not perfectly match the boundary in the real data. 
Now consider we build an additive feature attribute model $g(z)$ such as SHAP \cite{lundberg2017unified} given by Equation \ref{e:gz}:
\begin{equation}\label{e:gz}
    g(z) = \phi_{0} + \sum_{j=1}^{d} \phi_{j}(z_{j})
\end{equation}
where $z \in \{0,1\}^d$ is the projection of the input features from the original input space into binary values. The explanation model $g(z)$ models the classifier $f(x)$, attributing an effect $\phi_{j}$ to each feature and summing the effects of all feature attributions approximates the output $f(x)$. As per the additive model property {\it local accuracy} \cite{lundberg2017unified}, the result of Equation \ref{e:gz} is ${g(z)} \in Y$. Therefore, this model builds yet a third decision boundary, except that this is now local, {\it i.e.}, it models each instance. We construct a new prior $\Phi (X, Y)$ through constraining this third decision boundary with the boundary indicated by the real data using cross entropy loss (Equation \ref{e:cel}):
\begin{equation}\label{e:cel}
    \Phi (X, Y) = -\frac{1}{n}\sum_{i=1}^{n}y_i\log P (g(h_z(x_i))=y_i| {h_z(x_i)}),
\end{equation}
where we use a projection function $z = h_z(x)$ to represent $z$.
The intuition underlying our proposed approach is to use the XAI prediction for each training instance to provide a second loss as feedback to the training algorithm. The idea is that the new prior can help push the limits of data by learning more from less. With the original cross entropy loss (Equation \ref{e:ocel}):

\begin{equation}\label{e:ocel}
    \mathcal{L}(\theta;X,Y)   = -\frac{1}{n}\sum_{i=1}^{n}y_i\log P (f(x_i)=y_i| \bm{x_i ; \theta}), 
\end{equation}

 where $\theta$ are weights in the classifier. We construct the new objective function using the prior thus generating the new classifier we call the XAI-Augmented (XAIAUG) classifier (Equation \ref{e:new_obj_func}) :

\begin{equation}
\label{e:new_obj_func}
    \mathcal{L}_{XAIAUG}(\theta)  =  \mathcal{L}(\theta;X,Y) + \lambda  \Phi (X, Y)
\end{equation}
 
The expectation is that the classifications produced by the XAIAUG classifier, on average, have higher accuracy than the original classifier (which we will use as baseline).
It is also our expectation that the local accuracy of the additive feature attribution model computed for XAIAUG classifier is higher on average compared to the $Base$. The {\it local accuracy} is computed using Gradient SHAP \cite{erion2021improving} for each classifier at the end of every training epoch. 

The prior $\Phi (X, Y)$ needs to be differentiable so that we can backpropagate the error and update the weights $\theta$ of the classifier. In the next section, we will describe how we use Gradient SHAP to implement the prior.

\section{XAI-Augmented Approach}\label{XAIA}
In this section, we describe how we use Gradient SHAP and binary classification entropy (BCE) loss to implement the prior and then present the algorithm for training the XAIAUG classifier $f^{XAIAUG}(x)$.
From \cite{erion2021improving}, for each input feature $x_i$, we first obtain the feature attribution using Integrated Gradients and Expected Gradients:
\begin{equation}
\begin{aligned}
\phi_{(i)}(f,x) &= ExpectedGradient_j(x) \\
                 &=\int_{x'} IntegratedGradients_i(x,x') p_D(x')\delta x' \\
                  &= \int_{x'}(x_i - x_i')\int_{\alpha=0}^1\frac{\delta f(x'+\alpha \times (x-x'))}{\delta x_i}\delta \alpha \delta x'
\end{aligned}
\end{equation}

where $x'$ is a {\it baseline input feature} used as a starting point for calculating feature importance, which is drawn from a {\it background set} constructed with all training instances. $p_D(x')$ represents the probability distribution of the background dataset. Intuitively, the {\it baseline input feature} has the explanatory role as in representing the impact of the absence of each input feature on the prediction to contrast with the impact of each feature on the prediction when present in the input. By sampling the {\it baseline input feature} from the {\it background set}, we can measure the average impact of each feature. 

In our implementation, we solve the calculus by approximation, we have a dataset $D$, then we get:

\begin{equation}
     \phi_{(i)}(f,x)=\frac{1}{T}\sum_{x'\sim D}\sum_{k=1}^{m}\frac{\delta f(x' +\frac{k}{m}\times(x-x'))}{\delta x_i}
\end{equation}

where $D$ is the {\it background set} and $T$ is the sampling number. In Gradient SHAP, we have the assumption that all attribution go into features and do not consider $\phi_0$. Thus we have:

\begin{equation}\label{e:gh}
    g(h(x)) =  \sum_{j=1}^{d} \phi_{(i)}(f,x)
\end{equation}
In this way, we construct the differentiable prior and construct the new loss:

\begin{equation}\label{e:phi}
   \Phi (X, Y, \theta) = -\frac{1}{n}\sum_{i=1}^{n}y_i\log P (g(h_z(x_i))=y_i| {h_z(x_i)}; \theta)
\end{equation}

\begin{equation}\label{l_xaiaug}
    \mathcal{L}_{XAIAUG}(\theta)  =  \mathcal{L}(\theta;X,Y) + \lambda  \Phi (X, Y, \theta)
\end{equation} 

As Algorithm \ref{alg:xai_training} describes, we first randomly select 100 images from the training dataset as background for Gradient SHAP, for each iteration, we first apply gradient SHAP for each input image and get the sum of all SHAP values for each input image which is $\hat{\hat{y}}$. Then we construct SHAP loss using BCE loss to measure the difference between $\hat{\hat{y}}$ and their original labels. We update the classifier's weight by training with the sum of SHAP loss and classification loss.

\begin{algorithm}[!h]
    \caption{XAI-Augmented Training}
    \label{alg:xai_training}
    \textbf{Require}: GSHAP(GradientSHAP), BCELoss, Classifier $f$, BackProp, Background Dataset $D$, Total Training Epochs $e$\\
    \textbf{Input}:  Training frames $X_{train}$, Training Labels $Y_{train}$, Background frames Bframes\\
    \textbf{Parameter}: Weights $\theta$ of Classifier $f$\\
    \textbf{Output}: Updated weights $\theta$
    \begin{algorithmic}[1] 
        \STATE Let $p=0$.
        \STATE $Bframes \gets $ Extract 100 random frames from $D$
        \WHILE{$p<e$}       
            \STATE $\hat{y} \gets f(Tframes)$
            \STATE $SHAPval \gets GSHAP(Bframes,f,Tframes)$
            \STATE $\hat{\hat{y}} \gets SUM(SHAPval)$
            \STATE $PTCLoss \gets BCELoss(\hat{y} ,TLabels ) $
            \STATE $SHAPLoss \gets BCELoss(\hat{\hat{y}},TLabels)$
            \STATE $Loss \gets PTCLoss + SHAPLoss$
            \STATE BackProp and update weights.
            \STATE $p \gets p + 1 $
        \ENDWHILE
        \STATE \textbf{return} updated weight $\theta$
    \end{algorithmic}
\end{algorithm}

\section{Studies}\label{sec:studies}
In this section, we describe the methodology, metrics, and hypotheses for studies we conduct over three datasets. The methodology we investigate is to train the classifiers with the proposed XAIAUG approach described in Section \ref{XAIA} and compare against the baseline. The baseline refers to the classifiers described in Section \ref{base}. For all studies, we run our experiments through a 5-fold cross-validation and computed the averages for metrics across 5 folds, The choice of the number of epochs utilized for testing is based on the smallest loss obtained during training. The studies for PTX and COVID-19 are based on video frame blocks and not on the entire videos (See Section \ref{base} where we describe how frames are extracted from videos). Only when we discuss classification for ONSD dataset, we use videos rather than frames.

\paragraph{Datasets} Here we introduce the size of each dataset we used: PTX, ONSD and COVID-19. The first two data are not publicly available, collected under the DARPA POCUS-AI Program. The third dataset is publicly available. 

The PTX dataset contains 62 videos, 30 from patients with PTX, which is characterized by the absence of pleural line movement; and 32 from patients without PTX, where the pleural line movement is recognized. The average length of these videos is approximately three seconds at 20 frames per second with a convex probe at depths ranging from 4--12 cm. The ONSD dataset contains a total of 60 videos, 22 positive and 38 negative. The average length of the videos is 12.22 seconds long. 
The COVID-19 ultrasound dataset \cite{born2021accelerating} contains 202 videos, of which 97 are positive and 105 are negative. Each video has one of four labels: COVID-19, healthy, bacterial pneumonia, or viral pneumonia. Because our scope is limited to binary classification, we only use the data on healthy and COVID-19 classes. 

\paragraph{Metrics}
The studies utilize three performance metrics, which are used to build confusion matrices designative value of true positives (TP), true negatives (TN), false positives (FP), and false negatives (FN). The three performance metrics we use are Average Accuracy (AA), Balanced Accuracy (BA), and F1 score (F1). To compute all these, we need to compute sensitivity, specificity, and precision. See their formulas below:\\
\begin{gather}
   \noindent Average\ Accuracy = \frac{TP+TN}{TP+TN+FP+FN}\\
\newline
Precision = \frac{TP}{TP+FP}\\
\newline
Sensitivity = \frac{TP}{TP+FN}\\
\newline
Balanced \ Accuracy = \frac{Sensitivity + Specificity}{2}\\
\newline
Specificity = \frac{TN}{TN+FP}\\
\newline
F1 = \frac{2*Precision*Recall}{Precision+Recall} 
\end{gather}
 \\

Additionally, we compute local accuracy (LA) for Gradient
SHAP \cite{lundberg2017unified} to assess whether the additive feature attribution model correctly assigns the class of
each testing instance. LA \cite{lundberg2017unified} can be understood as a measure of local interpretability as it indicates how well the feature attributions model the instances. Note that in this paper we only present results for LA and do not demonstrate visualizations of interpretability.

The studies also utilize loss as a metric, which we plot across multiple epochs on training and testing data to determine the
influence of XAIAUG in decreasing overfitting. 

\paragraph{Hypotheses}
We adopt a general null hypothesis under which we compare the XAIAUG classifier against the baseline across the datasets for all metrics. In simple terms, considering each dataset as a variable $c \in C$, the metrics as $m \in M$, and a baseline $Base \in B$, we describe the general null hypothesis as:\\
$H_0$: The XAIAUG classifier produces lower or equal values for metric $m$ compared to the $Base$ for dataset $c$.

\section{Results and Discussion}\label{sec:results}
We present results for performance along selected metrics comparing the proposed approach with the baseline.

\subsection{Performance}\label{acc}
Tables \ref{tab:ptx-bas-xai}, \ref{tab:onsd-bas-xai}, and \ref{tab:covid-bas-xai} present the averages across all folds for AA, BA, and F1 for PTX, ONSD, and COVID-19. The results consistently show higher values for XAIAUG, particularly in the two metrics ({\it i.e.}, BA and F1) that account for data imbalance. These results reject the null hypotheses for the three datasets for BA and F1. For AA, the null hypothesis is rejected for the COVID-19 and ONSD datasets. On the other hand, the results for the COVID-19 data seem to be where the proposed XAIAUG approach produces the least improvement in the metric values. We observe that the COVID-19 dataset contains the largest amount of data with 202 videos, compared to the PTX dataset with 62 videos, and the ONSD dataset with 60 videos. This motivates us to investigate the reason for the least improvement in the COVID-19 dataset by analyzing the sensitivity of these metrics with respect to the data amounts in Section \ref{sensitivity}. 
\begin{table}[!h]
    \begin{tabular*}{\columnwidth}{l|cc|cc|cc}
        \hline
 {Metric} & 
        \multicolumn{2}{c}{AA} &
        \multicolumn{2}{c}{BA} &
        \multicolumn{2}{c}{F1} \\
    \hline
Method		&	Base	&	XAG	&	Base	&	XAG	&	Base	&	XAG	\\
    \hline
Ave	&	\textbf{0.807}	&	0.806	&	0.724	&	\textbf{0.781}	&	0.613	&	\textbf{0.693}	\\
    \hline
    \end{tabular*}
    \caption{The averages across all folds for AA, BA, and F1 for PTX dataset comparing XAIAUG (XAG) method against the baseline}
    \label{tab:ptx-bas-xai}
\end{table}

\begin{table}[!h]
    \begin{tabular*}{\columnwidth}{l|cc|cc|cc}
        \hline
 {Metrics} & 
        \multicolumn{2}{c}{AA} &
        \multicolumn{2}{c}{BA} &
        \multicolumn{2}{c}{F1} \\
    \hline
Method	 	&	Base	&	XAG&	Base	&	XAG&	Base	&	XAG	\\
    \hline
Ave	&	0.647	&	\textbf{0.690}	&	0.679	&	\textbf{0.740}	&	0.571	&	\textbf{0.637}	\\
    \hline

    \end{tabular*}
    \caption{The averages across all folds for AA, BA, and F1 for ONSD dataset comparing XAIAUG (XAG) method against Baseline}
    \label{tab:onsd-bas-xai}
\end{table}

\begin{table}[!h]
    \begin{tabular*}{\columnwidth}{l|cc|cc|cc}
        \hline
 {Metrics} & 
        \multicolumn{2}{c}{AA} &
        \multicolumn{2}{c}{BA} &
        \multicolumn{2}{c}{F1} \\
    \hline
Method		&	Base	&	XAG	&	Base	&	XAG	&	Base	&	XAG	\\
        \hline
Ave	&	0.663	&	\textbf{0.666}	&	0.636	&	\textbf{0.642}	&	0.554	&	\textbf{0.556}	\\
    \hline

    \end{tabular*}
    \caption{The averages across all folds for AA, BA, and F1 for the COVID-19 dataset comparing XAIAUG (XAG) method against the baseline }
    \label{tab:covid-bas-xai}
\end{table}

\begin{table*}[!t]
    \centering
    \begin{tabular}{l|ccc|ccc|ccc|ccc}
\hline
 {Data} & 
        \multicolumn{3}{c|}{COVID-19 dataset} &
        \multicolumn{3}{c|}{Small Set 1} &
        \multicolumn{3}{c|}{Small Set 2} &
        \multicolumn{3}{c}{Small Set 3} \\
\hline
Metric	&	AA	&	BA	&	F1	&	AA	&	BA	&	F1	&	AA	&	BA	&	F1	&	AA	&	BA	&	F1	\\
\hline
Base	&	0.663	&	0.636	&	0.554	&	0.543	&	0.500	&	0.291	&	0.718	&	0.709	&	0.528	&	0.648	&	0.643	&	0.558	\\
XAG	&	\textbf{0.666}	&	\textbf{0.642}	&	\textbf{0.556}	&	\textbf{0.592}	&	\textbf{0.572}	&	\textbf{0.457}	&	\textbf{0.741}	&	\textbf{0.765}	&	\textbf{0.637}	&	\textbf{0.659}	&	\textbf{0.652}	&	\textbf{0.585}	\\
\hline
\% diff	&	0.45\%	&	0.94\%	&	0.36\%	&	9.02\%	&	14.40\%		&57.04\%	&	3.20\%	&	7.02\%	&	20.64\%	&	1.70\%	&	1.39\%	&	4.48\%	\\
\hline
    \end{tabular}
    \caption{The averages across all folds for metrics AA, BA, F1, and percent difference comparing XAIAUG (XAG) method against the baseline for the entire COVID-19 dataset, and the smaller COVID-19 subsets Set 1, Set 2, and Set 3}
    \label{tab:covid-small-bas-xai}
\end{table*}


\subsection{Local Accuracy}
Table \ref{tab:la} shows the results for LA for the three datasets. The average LA of the XAIAUG classifier across all folds improves over the baseline for all three datasets. These results demonstrate that interpretability is improved since the additive feature attribution method is better aligned with the XAIAUG classifier than with the baseline. 
\begin{table}[!h]
    \begin{tabular*}{\columnwidth}{l|cc|cc|cc}
        \hline
 {Data} & 
        \multicolumn{2}{c}{PTX} &
        \multicolumn{2}{c}{ONSD} &
        \multicolumn{2}{c}{COVID-19} \\
    \hline
Method		&	Base	&	XAG	&	Base	&	XAG	&	Base	&	XAG	\\
\hline
Ave	&	0.725	&	\textbf{0.826}	&	0.827	&	\textbf{0.930}	&		0.897		&	\textbf{0.913}	\\
\hline
    \end{tabular*}
    \caption{The averages across all folds for LA obtained for PTX, ONSD, COVID-19 datasets comparing XAIAUG (XAG) method against the baseline }
    \label{tab:la}
\end{table}

\subsection{Sensitivity on Data Amounts}\label{sensitivity} 
We hypothesize that XAIAUG improves accuracy the least in the COVID-19 dataset because it is the one with the largest number of samples ({\it i.e.}, 2423 frames). Recall that our motivation is to learn more from the data when instances are limited. Because instances are not as limited in the COVID-19 dataset, the baseline classifier may have been able to learn generalization from the training dataset, potentially constraining XAIAUG's ability to significantly enhance accuracy.

To verify that the COVID-19 dataset is large enough and this is why XAIAUG does not improve accuracy as much as in the two datasets, we broke down the COVID-19 data into three smaller subsets we refer to as Small COVID-19 Set 1, Small COVID-19 Set 2, and Small COVID-19 Set 3 with 722, 907, and 794 frames, respectively. 

\paragraph{Hypothesis} We describe the general null hypothesis as $H_0$: For the COVID-19 dataset, the percentage performance differences for the XAIAUG classifier over the $Base$, calculated using \(\frac{XAIAUG - Base}{Base}\), for the metric $m \in M$, are less or equal when using only subsets versus using the entire dataset. 

To evaluate the hypothesis, we compare the percentage performance differences mentioned above between the XAIAUG and the baseline in Table \ref{tab:covid-small-bas-xai}.  

The differences for the subsets are larger compared to the whole dataset. While the difference for the entire COVID-19 dataset is below 1\% for all metrics, the differences for the small subsets are all above 1\%, varying from \%1.39 to \%57.04. The results across the three datasets reject the null hypothesis and validate that the potential for XAIAUG to improve a classifier's accuracy is constrained by dataset size.

\begin{table}[!h]
    \begin{tabular*}{\columnwidth}{l|cc|cc|cc}
        \hline
 {Metric} & 
        \multicolumn{2}{c}{AA} &
        \multicolumn{2}{c}{BA} &
        \multicolumn{2}{c}{F1} \\
    \hline
Method	&	L2	&	XAG	&	L2	&	XAG	&	L2	&	XAG	\\
\hline
Ave	&	\textbf{0.809}	&		0.806		&	0.726	&	\textbf{0.781}	&	0.622	&	\textbf{0.693}	\\
\hline
    \end{tabular*}
    \caption{The averages across all folds for AA, BA, and F1 for comparison between XAIAUG (XAG) and regularization for PTX}
    \label{tab:ptx-l2}
\end{table}

\begin{table}[!h]
    \begin{tabular*}{\columnwidth}{l|cc|cc|cc}
        \hline
 {Metric} & 
        \multicolumn{2}{c}{AA} &
        \multicolumn{2}{c}{BA} &
        \multicolumn{2}{c}{F1} \\ 
    \hline
Method &	L2	&	XAG	&	L2	&	XAG	&	L2	&	XAG	\\
\hline
Ave & \textbf{0.721} & 0.690 & \textbf{0.750} & 0.740 &\textbf{0.663} &0.637\\
\hline
    \end{tabular*}
    \caption{The averages across all folds for AA, BA and F1 for comparison between XAIAUG (XAG) and regularization for ONSD}
    \label{tab:onsd-l2}
\end{table}

\begin{table}[!h]
    \begin{tabular*}{\columnwidth}{l|cc|cc|cc}
        \hline
 {Metric} & 
        \multicolumn{2}{c}{AA} &
        \multicolumn{2}{c}{BA} &
        \multicolumn{2}{c}{F1} \\
    \hline
Method	   &   L2  &   XAG   &   L2 &    XAG   &   L2  &   XAG  \\
\hline
Ave & 0.635 & \textbf{0.666} & 0.609 & \textbf{0.642} & 0.486 & \textbf{0.556}\\
\hline
    \end{tabular*}
    \caption{The averages across all folds for AA, BA and F1 for comparison between XAIAUG (XAG) and regularization for COVID-19}
    \label{tab:covid-l2}
\end{table}

\subsection{Comparison against Regularization}\label{Regularization} 
Tables \ref{tab:ptx-l2}, \ref{tab:onsd-l2}, and \ref{tab:covid-l2} show results for the previously defined performance metrics but now comparing the XAIAUG classifier with a different baseline -- L2. The compared classifier is different from the original baseline in that the L2 norm is added to the loss function for regularization. As when against the baseline, in general, XAIAUG produces better performance in metrics that consider class imbalance. The exception is in BA for the ONSD dataset. We next look at the loss values plotted along training epochs. 

\begin{figure*}[!h]
    \centering
    \includegraphics[clip=true, width=0.90\textwidth]{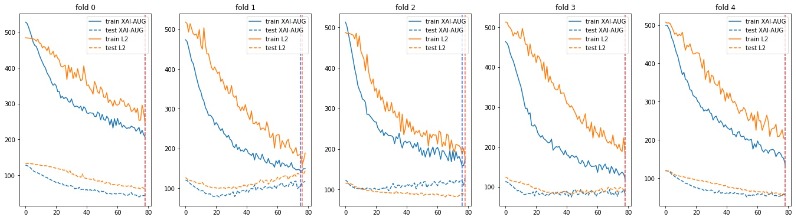}
    \caption{Loss (y-axis) and epochs (x-axis) comparing XAIAUG to L2 for PTX.}
    \label{fig:ab}
\end{figure*}
\begin{figure*}[!h]
    \centering
    \includegraphics[clip=true, width=0.90\textwidth]{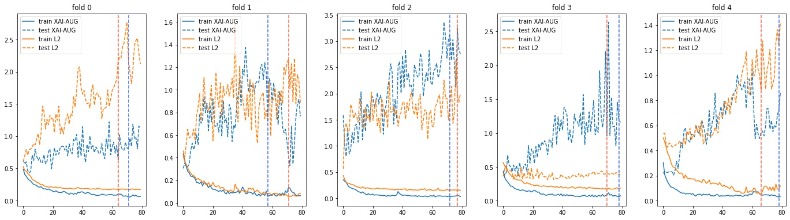}
    \caption{Loss (y-axis) and epochs (x-axis) comparing XAIAUG to L2 for ONSD.}
    \label{fig:bc}
\end{figure*}
\begin{figure*}[!h]
    \centering
    \includegraphics[clip=true, width=0.90\textwidth]{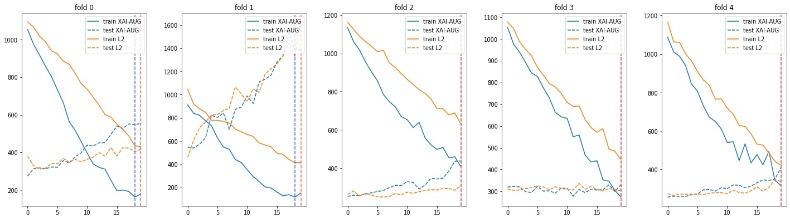}
    \caption{Loss (y-axis) and epochs (x-axis) comparing XAIAUG to L2 for COVID-19.}
    \label{fig:cd}
\end{figure*}

Figures \ref{fig:ab}, \ref{fig:bc}, and \ref{fig:cd} depict the loss with respect to epochs for the proposed approach and the alternate baseline L2 with L2 regularization classifiers for the three datasets. The vertical lines outline the epoch with minimum training loss. The curves correspond to training and testing data so we can analyze overfitting. Analogous to improvements in accuracy, the plots reveal that the XAIAUG classifier often produces values for loss lower than L2 in many folds but not for all. In Figure \ref{fig:ab}, the XAIAUG testing data shows loss values at the latest epochs to be lower than those from L2 testing data in four out of the five folds, suggesting a potential decrease in overfitting. In ONSD data (Figure \ref{fig:bc}), the decrease in overfitting happens in three out of five folds. In Figure \ref{fig:cd}, the XAIAUG testing data shows loss values at the latest epochs to be higher than those from L2 in folds zero and two. In folds one, three, and four, it goes up at the very end. 

The conclusion that can be drawn is that the improvements in accuracy are consistent with the reduced overfitting. This suggests the reduction in overfitting could be the reason for the increments in accuracy. In future work, we shall investigate the conditions under which our approach is useful.

\section{Conclusions}\label{sec:conclusion}
This paper proposes an XAI-Augmented approach to train classifiers with scarce data, with the aim to learn more from less. To investigate and analyze the proposed approach in terms of accuracy, we utilize three ultrasound video datasets, PTX, ONSD, and COVID-19. We present studies to better comprehend why and when this approach improves performance. We found a consistent improvement in Balanced Accuracy and F1 for all three datasets. The study of subsets of COVID-19 datasets shows that our approach is sensitive to data amounts and has the potential for improving classifiers when training data is scarce. By comparing against regularization, we found that the improvements in accuracy are consistent with reduced overfitting. Overall, our results are consistent and conclusions are drawn based on results without major variations.

\section{Limitations and Future Work}\label{sec:limitation}
This work is limited in that it does not investigate multiple approaches that have been used to increase accuracy such as data augmentation, single-class learning, and zero-, one-, or few-shot learning. Future studies should demonstrate how the proposed XAIAUG approach compares to these other ways to improve accuracy when only small data amounts are available to train a classifier. In future work, synthetic datasets may be used to determine what data conditions, if any, cause these variations. The other alternative is to investigate the impact of the XAIAUG approach on aspects such as incremental learning \cite{shmelkov2017incremental}.

\subsubsection{Ethical Statement}
The authors have conducted this work after receiving approval from IRB in their respective institutions and from the funding agency sponsoring this work.
\subsubsection{Acknowledgements}
This research was developed with funding from the Defense Advanced Research Projects Agency (DARPA) under the POCUS AI Program (award no. HR00112190076). The views, opinions and/or findings expressed are those of the authors and should not be interpreted as representing the official views of the Department of Defense or the U.S. Government.

\bibliographystyle{named}
\bibliography{ijcai24}

\end{document}